%% file: main.tex
\documentclass[10pt,twocolumn,letterpaper]{article}

\usepackage[pagenumbers]{cvpr} %

\input{preamble}

\definecolor{cvprblue}{rgb}{0.21,0.49,0.74}
\usepackage[pagebackref,breaklinks,colorlinks,allcolors=cvprblue]{hyperref}

\def\paperID{*****} %
\def\confName{CVPR}
\def\confYear{2026}

\newcommand{\LEGO}{LEGO\textsuperscript{®}}

\title{BrickWright: Agentic Creation of Physically Feasible Brick Builds}

\author{Ava Pun\textsuperscript{1}\quad
Kevin Dai\textsuperscript{1}\quad
Sally Chen\textsuperscript{1}\quad
Maxwell Jones\textsuperscript{1}\quad
Ruixuan Liu\textsuperscript{1}\\
Kangle Deng\textsuperscript{2}\quad
Changliu Liu\textsuperscript{1}\quad
Maneesh Agrawala\textsuperscript{3}\quad
Deva Ramanan\textsuperscript{1}\quad
Jun-Yan Zhu\textsuperscript{1}\\
\textsuperscript{1}Carnegie Mellon University\quad
\textsuperscript{2}Independent Researcher\quad
\textsuperscript{3}Stanford University
}

\begin{document}
\makeatletter
\g@addto@macro\@maketitle{\input{fig/teaser}}
\makeatother
\maketitle
\input{sec/0_abstract}    
\input{sec/1_intro}

\input{sec/2_related_work}
\input{sec/4_method}
\input{sec/5_experiments}

\input{sec/9_conclusion}
\clearpage
{
    \small
    \bibliographystyle{ieeenat_fullname}
    \bibliography{main}
}

\input{sec/X_suppl}

\end{document}

%% file: preamble.tex
\usepackage{microtype}

\renewcommand{\paragraph}[1]{\vspace{.5em}\noindent\textbf{#1}}

\usepackage{makecell}
\usepackage{multirow}

\usepackage{listings}
\definecolor{codekeyword}{HTML}{A65301} %
\definecolor{codestring}{HTML}{206873}  %
\definecolor{codenumber}{HTML}{873461}  %
\definecolor{codecomment}{HTML}{2D5F2C} %
\lstdefinestyle{bricklayer}{
    language=Python,
    basicstyle=\ttfamily\scriptsize,
    keywordstyle=\color{codekeyword}\bfseries,
    stringstyle=\color{codestring},
    numberstyle=\color{codenumber},
    commentstyle=\color{codecomment}\itshape,
    breaklines=true,
    breakatwhitespace=true,
    frame=single,
    captionpos=b,
}
\makeatletter
\newcommand{\lstdivider}{\makebox[0pt][l]{\hspace*{-\lst@frametextsep}\rule[0.2em]{\dimexpr\linewidth+\lst@frametextsep+\lst@frametextsep\relax}{\lst@framerulewidth}}}
\newcommand{\lstcornerimage}[2][0.25\linewidth]{\makebox[0pt][l]{\hspace*{\linewidth}\llap{\raisebox{-\dp\strutbox}[0pt][0pt]{\setlength{\fboxsep}{0pt}\setlength{\fboxrule}{\lst@framerulewidth}\fbox{\includegraphics[width=#1]{#2}}}}}}
\makeatother

%% file: fig/teaser.tex
\begin{minipage}{\textwidth}
  \centering
  \includegraphics[width=\linewidth]{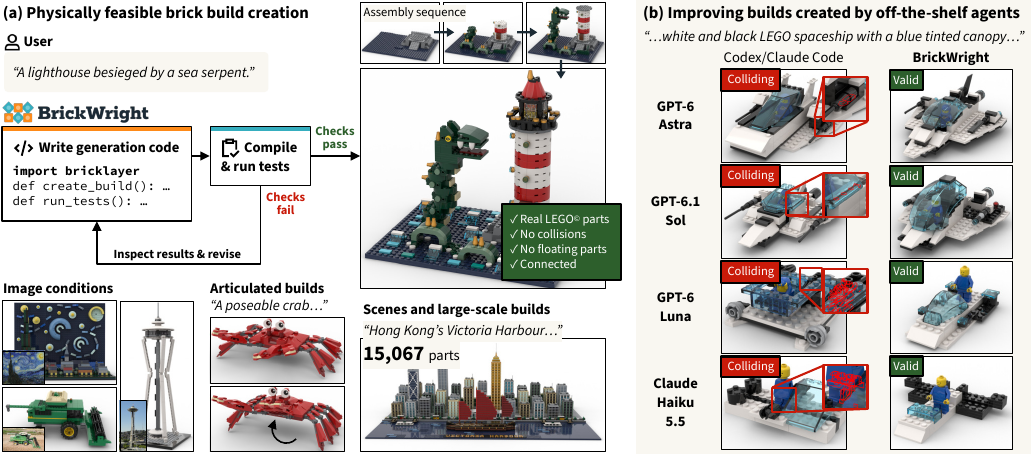}
  \captionof{figure}{\textbf{BrickWright overview.} BrickWright instructs pre-trained LLM agents to generate physically feasible brick builds by writing code using a Python SDK, BrickLayer, in a plan-execute-validate loop utilizing BrickLayer's physical validation tools (a). BrickWright accepts text or image inputs and can produce articulated and large-scale builds, all physically valid. Agents equipped with BrickWright produce higher-quality builds with fewer physical violations (b). Red outlines in insets indicate mesh intersections.}
  \label{fig:teaser}
\end{minipage}
\par\vspace{1em}

%% file: sec/0_abstract.tex
\begin{abstract}
Frontier coding agents are becoming increasingly adept at 3D modeling. Yet, they still face challenges when building objects meant to be assembled in the physical world, often creating invalid assemblies with floating, colliding, or disconnected parts. We argue that providing agents with tools for assembling builds and verifying their physical feasibility is sufficient for them to produce physically valid assemblies that still adhere to input prompts. Focusing on the domain of \LEGO{} brick builds, we present BrickWright, a framework for generating physically feasible brick builds from text or image prompts using pre-trained agents, where a build is physically feasible if it uses only real parts, has no colliding or floating parts, and forms a connected structure if possible. BrickWright guides agents to create builds by writing code with BrickLayer, a Python toolkit with two main components: an ergonomic brick assembly API that halves inference cost compared to working in raw LDraw, and a validator that provides feedback on physical feasibility. Whereas baseline agents achieve physical validity rates ranging from only 9\% (GPT-6 Luna) to 69\% (GPT-6 Astra), BrickWright-guided agents achieve 100\% validity on a sample of 512 prompts, while improving prompt adherence for lightweight agents and maintaining it for stronger agents. Additionally, they outperform specialized models on the text-to-brick task. Code is available on our project page: \texttt{\href{https://avalovelace1.github.io/brickwright/}{avalovelace1.github.io/brickwright}}.
\end{abstract}

%% file: sec/1_intro.tex
\section{Introduction}
\label{sec:intro}

Large language model (LLM) agents are becoming increasingly adept at 3D modeling. Without additional fine-tuning, commercially available agents can successfully produce articulated objects~\cite{zhou2026articraft, wang2026adsl}, scenes~\cite{yin2026viga}, and constructive solid geometry models~\cite{lin2026procedura}.
However, they still face challenges when building objects for the physical world. We focus on the task of creating \textit{multi-part assemblies}: real-world objects made from a set of prefabricated parts joined by fixed connectors, such as \LEGO{} builds. Multi-part assemblies must obey multiple constraints: all parts must be drawn from a preset library, parts must not collide or float, and the joined parts typically form a single connected component.

We argue that designing a \textit{toolkit} for multi-part assembly creation is sufficient to enable agents to create physically valid builds. This toolkit should include (1) an \textit{assembly API}, which provides an ergonomic programmatic interface for 3D creation; and (2) a \textit{validator}, which provides feedback on a finished build's physical validity. The need for a toolkit is \textit{independent of how powerful a coding agent is}. We observe that GPT-6 Astra with Codex, when asked to make brick builds, already writes scripts to generate and validate its output more than 80\% of the time~(\cref{sec:vanilla-analysis}). For complex physical domains, this self-validation is often not enough, resulting in builds that do not satisfy physical constraints, containing colliding, floating, or disconnected parts~(\cref{fig:physical-failure-cases}). Even if an agent is powerful enough to write a perfect toolkit, separately developing a toolkit remains important to reduce token usage and avoid repeated work. Existing toolkits for agentic 3D generation~\cite{zhou2026articraft, wang2026adsl} are usually intended for digital asset creation; as they do not consider physical constraints, they cannot be repurposed as-is to create real-world objects.

\input{fig/physical-failure-cases}

We focus on the domain of \LEGO{} generation as it is a commonly available yet complex system, consisting of thousands of parts that connect in dozens of ways. Our key insight is to merge agentic code-as-3D-models with the regularity of brick parts. Parts can be modeled as geometries with localized connectors that have \textit{programmatic rules} for assembly (e.g., studs interlock with antistuds). From this perspective, we leverage the physical production rules for snap-fit brick builds~\cite{christiansen1961brick} to create an ergonomic assembly API. Then, given an image or text prompt, our task is to generate a program using this API that, when compiled, produces a \LEGO{} brick build in the standard LDraw format~\cite{ldrawdocumentation}. This build should include a list of library part placements, and it must be \textit{physically feasible}, having no intersecting or floating parts and forming a connected structure if possible. While there are several specialized brick generation methods~\cite{xu2025legoace, ge2024microbuildings, ge2025legomaker, kulits2026bricknet, pun2025brickgpt, xu2026stable, ni2026brickanything, yuan2026pvpo}, some of which consider physical constraints, they generally exhibit worse generalization and lower fidelity than commercial LLM agents on the same task. Hence, we wish to achieve this task agentically.

Toward this goal, we introduce \textit{BrickLayer}, a Python package for assembling and validating brick builds, alongside \textit{BrickWright}, a framework that guides LLM agents through creating validated brick builds with BrickLayer. Agents equipped with BrickWright create higher-quality builds with fewer physical violations; on a test set of 512 prompts, all agents---including lightweight models like GPT-6 Luna and Claude Haiku 5.5---achieve a 100\% validity score without a significant cost increase or drop in fidelity. BrickWright is especially useful for lightweight models: it improves Luna's performance from 9\% to 100\% valid builds and boosts its prompt adherence at 2\% the price of frontier models. BrickWright additionally outperforms specialized brick generation models on validity and fidelity. Code is available on our project page: \texttt{\href{https://avalovelace1.github.io/brickwright/}{avalovelace1.github.io/brickwright}}.

%% file: fig/physical-failure-cases.tex
\begin{figure}[t]
  \centering
  \includegraphics[width=\columnwidth]{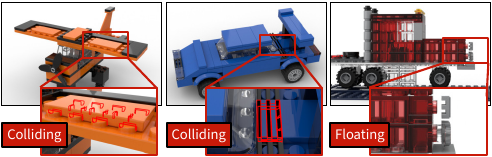}
  \caption{\textbf{Physical violations in builds by Codex (GPT-6 Astra, high reasoning).} Red outlines in insets show mesh intersections.}
  \label{fig:physical-failure-cases}
\end{figure}

%% file: sec/2_related_work.tex
\input{fig/method}

\section{Related work}
\paragraph{Brick Build Generation.} Early work treats brick build generation as an optimization problem over a target shape. Legolization \cite{luo2015legolization} voxelizes a 3D mesh, merges voxels into standard bricks, and iteratively re-merges regions identified as unstable. Subsequent work captures finer geometric detail through feature-aware shape approximation and non-rectangular bricks \cite{zhou2019vivid, zhou2023sketchart}, assembles bricks sequentially via Bayesian optimization \cite{kim2020combinatorial}, or optimizes Technic assemblies to match annotated sketches~\cite{xu2019technic}. Retrieval-based methods map images to figurines assembled from prebuilt templates \cite{zhou2022brickyourself, ge2024figurines}. These methods are limited by small brick vocabularies, explicit geometric guidance, or fixed templates.

Learning-based methods generate brick structures directly from data. These approaches adapt graph generative models \cite{thompson2020dgmg}, predict voxel representations that are assembled into bricks algorithmically \cite{lennon2021image2lego, chen2026resembrick}, learn sequential construction policies~\cite{chung2021brickbybrick}, or use diffusion
to predict labeled voxels within a single category \cite{ge2024microbuildings}. Autoregressive methods generate bricks conditioned on text or
images. BrickGPT \cite{pun2025brickgpt} fine-tunes an LLM to generate stable structures, but is restricted to eight rectangular brick types across 21 object categories. LEGO-Maker \cite{ge2025legomaker} generates builds from images, but
supports only axis-aligned orientations and two categories. LegoACE \cite{xu2025legoace} and BrickNet \cite{kulits2026bricknet} scale to open-vocabulary generation by training on
large collections of human-designed builds. However,
LegoACE learns connectivity implicitly and can produce disconnected components, while BrickNet represents assemblies as connector graphs but accumulates collisions over longer build sequences. Brick-Composer \cite{liu2026brickcomposer} instead approaches assembly from target views as sequential brick selection and pose estimation, training a multimodal LLM on simulator-generated data and feedback. Other methods improve buildability through preference optimization or reinforcement learning~\cite{ni2026brickanything, xu2026stable}. In contrast, BrickWright utilizes pre-trained models to build arbitrary objects without task-specific training, verifies collisions and connectivity in a feedback loop, and supports articulation. Concurrent with our work, BrickBench \cite{kulits2026brickbench} introduces a toolkit for agentic brick build creation, focusing on benchmarking the performance of different agents.

\paragraph{Agentic 3D Generation.} LLM agents have been used successfully to create 3D scenes. Recent works use a plan-execute-critique loop \cite{hu2024scenecraft, ling2025scenethesis} or reconstruct scenes from images through iterative code execution and visual inspection \cite{yin2026viga, he2026seig}. Other works fine-tune LLMs to reason directly over 3D spatial inputs \cite{wan2026nala} or to generate tool-call trajectories \cite{he2026sceneorchestra}. These methods place objects in continuous space and rely on optimization or visual judgment to resolve collisions and floating objects. In BrickWright, explicit connector rules enable deterministic checks of attachment validity.

Representing shapes as programs provides structured descriptions of geometry and part relationships \cite{sharma2018csgnet, tian2019shapeprograms, jones2020shapeassembly}. Separately,
LLMs have been shown to reason more reliably when they write code and offload exact
computation to an interpreter \cite{gao2023pal}. Building on both ideas, agents have been applied to modeling individual objects as code, typically through plan-execute-critique loops. Prior work augments this loop with API retrieval \cite{lu2025ll3m}, domain-specific languages \cite{wang2026adsl}, or procedural part-based modeling \cite{lin2026procedura}, with some supporting articulation \cite{wang2026adsl, lin2026procedura, zhou2026articraft, le2025articulateanything}. In CAD, LLMs have been
fine-tuned to generate parametric modeling code \cite{xie2025text2cadquery, guan2025cadcoder,
kim2026brepcoder}, while tool-augmented agents generate and refine parts and assemblies~\cite{mallis2025cadassistant,shui2026articad,adler2026aadvark,zheng2026spada}. Articraft \cite{zhou2026articraft} generates articulated assets using a custom SDK and a minimal harness that provides textual feedback from tools that probe the model code. Unlike these systems, where the agent generates geometry or infers inter-part relationships, BrickWright assembles fixed parts whose connectors determine how they attach and move. 

\paragraph{Physics-Aware Generation.} To improve physical validity, prior work refines shapes or layouts after generation \cite{chen2024atlas3d, xu2025fun3d, zhan2025physiopt, ling2025scenethesis, li2025phipg}, fine-tunes generators on simulator feedback \cite{li2025dso}, or guides diffusion sampling with physical constraints \cite{luo2025physpart}. Geometric analysis has also been used to plan LEGO assembly sequences without explicit connector information~\cite{agrawala2003assembly}. For brick structures, stability has been assessed through force-based equilibrium analysis \cite{luo2015legolization, pun2025brickgpt, liu2024stablelego} or physics simulation \cite{kim2020combinatorial, wen2026bricksim}. In contrast, BrickWright uses an executable verifier to check for collisions, floating parts, and connectivity, allowing a pre-trained agent to iteratively repair builds without task-specific training. Incorporating force-based stability analysis~\cite{luo2015legolization,liu2024stablelego} is a natural extension.
\label{sec:related_work}

%% file: fig/method.tex
\begin{figure*}[t]
    \centering
    \includegraphics[width=\textwidth]{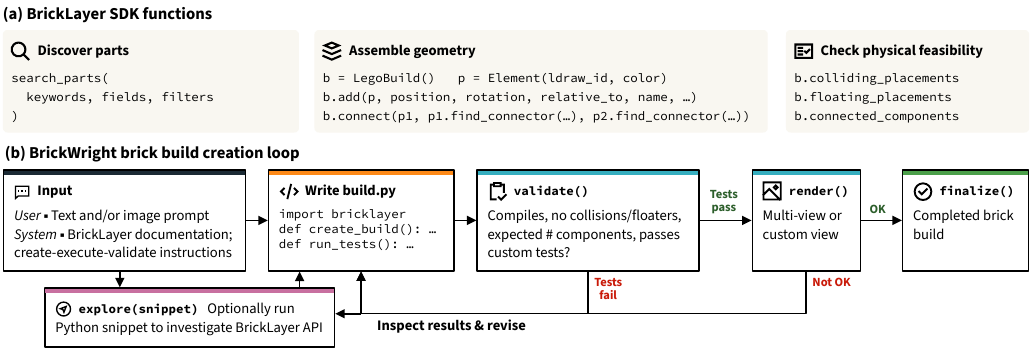}
    \caption{\textbf{Method overview.} (a) BrickLayer supports catalog discovery, compositional construction with typed connectors, and physical validation. (b) BrickWright prompts a pre-trained LLM agent to write and revise \texttt{build.py}, which contains brick build creation code and agent-defined tests. The \texttt{validate} tool runs fixed collision and floating checks alongside the agent's tests, while the \texttt{render} tool provides views for visual inspection. The agent calls \texttt{finalize} once all checks pass. The agent may optionally run \texttt{explore} at any time to investigate the BrickLayer API and written brick build. Arrows summarize information flow rather than a fixed tool-call sequence.}
    \label{fig:method}
\end{figure*}

%% file: sec/4_method.tex
\section{BrickWright}
\label{sec:method}

Given a natural-language or image description, our task is to generate a \textit{physically feasible brick build} from a fixed part catalog. A brick build is defined as an ordered sequence of \textit{placements}. Each placement consists of an \textit{element} or a nested brick build, together with a rigid transformation relative to the containing build. An element specifies a single \textit{part} drawn from the catalog and a color. Elements expose typed \textit{connectors}, such as studs, whose compatibility and relative alignment determine valid \textit{connections}. A build is \textit{valid} if it uses only catalog parts and contains no colliding (intersecting) or floating parts. It forms a \textit{connected component} if one can draw a path of connections between any two elements in the build. We define a build to be \textit{physically feasible} if it is valid and contains as few connected components as possible. More complex feasibility considerations, such as ensuring a build is stable and has an executable assembly sequence, are left to future work. Note that although having a single connected component is desirable as it makes a build less likely to break apart in the physical world, builds with multiple components are still necessary in some cases: for example, two separate vehicles in a car chase scene.

We introduce \textit{BrickWright} (\cref{fig:method}), a framework for generating physically feasible brick builds using LLM agents.
Using a lightweight harness, BrickWright guides a single agent to write Python code that produces a brick build using a custom toolkit called \textit{BrickLayer}. If the build passes a set of predefined and agent-written tests, it is accepted; otherwise, the agent makes edits informed by the test results, following the commonly used plan-execute-validate loop.

\subsection{BrickLayer}

LDraw~\cite{ldrawdocumentation}, the standard text format for \LEGO{} builds, includes no physical validation and is not very \textit{ergonomic}: it only allows users to place parts and sub-models by their absolute 3D poses. This makes it difficult to mate parts by their connectors or place them relative to others, which can lead to wasted tokens and misplaced components. To provide agents with an ergonomic interface for creating and validating brick builds, we introduce \textit{BrickLayer}~(\cref{lst:bricklayer}), a Python package that includes functions for part discovery, assembly, and validation. BrickLayer provides a unified programmatic representation for brick builds, enabling them to be easily understood and edited by humans and agents; builds can be translated from BrickLayer to LDraw and vice versa. See \cref{sec:system-prompts} for BrickLayer documentation.

\input{lst/bricklayer}

\input{tab/main-quantitative}

\paragraph{Discovery.} BrickLayer allows users to search the LDraw catalog~\cite{ldrawlibrary}, which contains 17k+ parts. This search accepts keywords matched against the part's official LDraw name. Results are ordered by their BM25 score~\cite{robertson2009bm25} multiplied by a popularity factor that scales logarithmically with the number of official \LEGO{} sets containing the part. LDraw names are parsed into fields such as category, subcategory, and dimensions; users may target specific fields with their keyword search and filter results using exact field matches.

\paragraph{Assembly.} The central objects of BrickLayer are \texttt{Element}, representing a single colored part, and \texttt{LegoBuild}, representing a brick build. Both objects support a unified \texttt{LegoGeometry} interface; this interface exposes dimensions and available connectors, which can be filtered by type and position. We use BrickNet's connector annotations~\cite{kulits2026bricknet}, which mark the local frames of the connectors on each part, along with compatible connector types and the joints they form: fixed, revolute, ball, prismatic, or cylindrical. For example, studs and antistuds join with a revolute joint. As BrickNet annotations do not differentiate between prismatic and cylindrical joints, we determine a sliding joint's type by rotating it $15^\circ$ about its axis in one direction; if a collision occurs, we mark the joint prismatic.

\texttt{LegoBuild}s have a recursive structure: they may contain both \texttt{Element}s and other \texttt{LegoBuild}s, allowing complex models to be built from smaller components. Placements can be made at arbitrary poses, fixed to a grid, or by mating specified connectors. Users may name placements, allowing them to place components relative to or directly target the connectors of a named component. Users may also explicitly mark movable components with articulation limits corresponding to their joint type.

For convenience, BrickLayer includes several high-level construction functions, introduced after noting repeated patterns in agent-generated code. These include \texttt{brick}, \texttt{plate}, \texttt{tile}, \texttt{slope}, and \texttt{inverted\_slope} to instantiate common parts by specifying their dimensions; \texttt{minifigure}, \texttt{door}, and \texttt{window} to create common multi-component sub-builds; \texttt{mirror} to reflect a build along a given axis; and \texttt{fill\_box}, \texttt{fill\_slope}, and \texttt{fill\_inverted\_slope} to fill a cuboid or wedge in a manner guaranteed to be valid and connected if possible.

\paragraph{Validation.} A \texttt{LegoBuild} provides methods to list collisions, floating elements, and disconnected components. To determine whether two elements collide, we inset each mesh by 0.1 mm before checking for intersections, following BrickNet; snap connections between bricks are so tight that a naive mesh intersection check would report a false positive. The floater check first translates a build so its lowest point is at the ground plane. Then, it marks an element \textit{floating} if it cannot reach a grounded element through mated connectors or \textit{supporting surface contacts}. A surface contact is supporting if it has an upward component; lateral contact alone is insufficient. Two connectors form a connection if their local frames align, depending on their joint type; for example, a fixed joint requires the two frames to align exactly, while a revolute joint additionally admits a rotational degree of freedom. For builds with user-annotated articulations, a \texttt{posed} method allows an agent to create a snapshot of the build with joints set to specified articulation parameters, to validate physical feasibility under joint motion. Posing works with models whose kinematic structure forms a tree.

\input{fig/ablation}

\input{fig/extensions}

\subsection{Building with BrickLayer}

BrickWright is a framework for agentic brick build creation using BrickLayer. It provides a single agent with a set of tools and prompts it to create a build by writing and revising a single Python file in a plan-execute-validate loop, inspired by Articraft~\cite{zhou2026articraft}. The BrickWright environment exposes the tools \texttt{validate}, which runs agent-written tests alongside fixed collision and floating checks; \texttt{render}, which renders the build into a multi-view sheet or from a single specified view; and \texttt{finalize}, which marks the build as finished.

In its system prompt, BrickWright instructs an agent to make a short internal plan, then write a Python file, \texttt{build.py}, containing two functions: \texttt{create\_build()}, which returns a \texttt{LegoBuild}, and \texttt{run\_tests()}, which runs a set of agent-written tests against the completed build.
After creating \texttt{build.py}, the agent is instructed to iteratively call \texttt{validate} and edit the build file until all tests pass, then call \texttt{render}; if it judges the render satisfactory, it should then call \texttt{finalize}; otherwise, it should revise \texttt{build.py} and repeat validation. Besides these instructions, BrickWright includes only the BrickLayer documentation in its system prompt, and the input text or image in a user prompt. It does not provide the agent with detailed planning advice, complete build examples, or building tips beyond the plan-execute-validate loop; the loop alone is sufficient for even lightweight agents to produce a feasible build.

BrickWright can be used alongside commercial frameworks such as Codex, which allows general file access and script execution. However, to demonstrate its utility alone, we provide the agent with only two additional tools during experiments: \texttt{read} / \texttt{write} / \texttt{insert} / \texttt{edit\_build} to read and modify \texttt{build.py}, and \texttt{explore} to run a Python snippet with BrickLayer and the written \texttt{LegoBuild} in scope, returning its output as text.

%% file: lst/bricklayer.tex
\begin{lstlisting}[
    style=bricklayer,
    float=t,
    caption={\textbf{BrickLayer usage.} Users can add elements, join them via selected connectors, and assess feasibility. The BrickLayer script is easier to understand than its corresponding LDraw output.},
    label={lst:bricklayer},
    escapeinside={(*@}{@*)}
]
b = LegoBuild() # Initialize build
socket = b.add(Element("14418", Color.ORANGE))
ball = Element("3170", Color.MEDIUM_AZURE)
b.connect( # Search for and mate compatible connectors
  ball,
  ball.find_connector("towball", toward=Direction.RIGHT),
  to=socket.find_connector("towball_socket")
)
assert (not b.colliding_placements # Check feasibility
        and not b.floating_placements
        and len(b.connected_components == 1)
write_ldraw(b, "build.ldr") # Export to LDraw
(*@\lstdivider@*)
Output LDraw:
1 25 20 -8 10 1 0 0 0 1 0 0 0 1 14418.dat
(*@\lstcornerimage{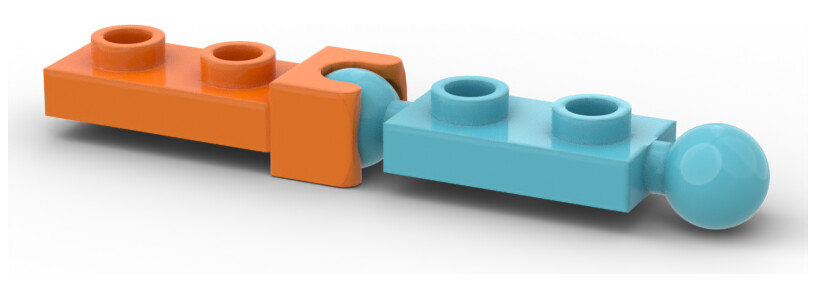}@*)1 322 80 -8 10 1 0 0 0 1 0 0 0 1 3170.dat
\end{lstlisting}

%% file: tab/main-quantitative.tex
\begin{table*}[t]
    \caption{\textbf{Quantitative results on BrickNet-512.} Valid, No col., and No fl. are valid, collision-free, and floater-free build rates. Comp., Parts, and Unique are mean component, part, and distinct part-ID counts. PE: Perception Encoder. SL2: SigLIP 2. Checks: mean \texttt{validate} calls. Bold/underline mark best/second-best results within each agent group. Costs use October 2026 token pricing, including cache savings.}
    \label{tab:main-quantitative}
    \centering
    \setlength{\tabcolsep}{2pt}
    \begin{tabular*}{\textwidth}{@{}cl@{\hspace{3pt}\extracolsep{\fill}}*{12}{c}@{}}
        \toprule
        \multicolumn{2}{@{}l}{Method} & Valid$\uparrow$ & No col.$\uparrow$ & No fl.$\uparrow$ & Comp.$\downarrow$ & VQA$\uparrow$ & PE$\uparrow$ & SL2$\uparrow$ & Parts & Unique & Checks$\downarrow$ & Time$\downarrow$ & Cost$\downarrow$ \\
        \midrule
        \multicolumn{2}{@{}l}{BrickGPT~\cite{pun2025brickgpt}} & 0.990 & 1.000 & 0.990 & 2.67 & 0.0658 & 0.1547 & 0.0514 & 144.8 & 7.1 & --- & 247s & --- \\
        \multicolumn{2}{@{}l}{BrickNet~\cite{kulits2026bricknet}} & 0.318 & 0.320 & 0.943 & 1.03 & 0.6077 & 0.2846 & 0.6407 & 31.8 & 11.5 & --- & 5s & --- \\
        \midrule
        \multirow{4}{*}{\rotatebox[origin=c]{90}{Haiku 5.5}} & Claude Code & 0.328 & 0.428 & 0.387 & 3.59 & 0.6032 & 0.2963 & 0.7286 & 12.8 & 5.3 & --- & 168s & \$0.04 \\
         & Ours w/o assembly API & \underline{0.822} & \underline{0.889} & \underline{0.828} & \underline{2.07} & 0.5744 & 0.2940 & 0.7009 & 16.6 & 4.8 & \underline{9.29} & 279s & \$0.13 \\
         & Ours w/o validator & 0.586 & 0.744 & 0.775 & 2.99 & \underline{0.6390} & \underline{0.3012} & \underline{0.7481} & 18.3 & 7.4 & --- & 119s & \$0.03 \\
         & \textbf{Ours (full)} & \textbf{1.000} & \textbf{1.000} & \textbf{1.000} & \textbf{1.00} & \textbf{0.6523} & \textbf{0.3056} & \textbf{0.7660} & 21.5 & 7.6 & \textbf{2.89} & 143s & \$0.05 \\
        \midrule
        \multirow{4}{*}{\rotatebox[origin=c]{90}{6 Luna}} & Codex & 0.086 & 0.131 & 0.135 & 10.11 & 0.6504 & 0.3025 & 0.7593 & 21.1 & 6.1 & --- & 150s & \$0.02 \\
         & Ours w/o assembly API & \underline{0.996} & \textbf{1.000} & \underline{0.996} & \underline{1.04} & 0.3342 & 0.2586 & 0.4972 & 7.2 & 2.0 & \underline{15.25} & 405s & \$0.06 \\
         & Ours w/o validator & 0.529 & \underline{0.561} & 0.840 & 2.03 & \textbf{0.6975} & \textbf{0.3111} & \textbf{0.7920} & 28.5 & 8.5 & --- & 510s & \$0.03 \\
         & \textbf{Ours (full)} & \textbf{1.000} & \textbf{1.000} & \textbf{1.000} & \textbf{1.00} & \underline{0.6945} & \underline{0.3105} & \underline{0.7766} & 27.4 & 7.8 & \textbf{4.62} & 415s & \$0.03 \\
        \midrule
        \multirow{4}{*}{\rotatebox[origin=c]{90}{6.1 Sol}} & Codex & 0.641 & 0.654 & 0.781 & 2.00 & 0.7610 & \underline{0.3241} & \textbf{0.8700} & 75.9 & 15.8 & --- & 346s & \$0.29 \\
         & Ours w/o assembly API & \textbf{1.000} & \textbf{1.000} & \textbf{1.000} & \underline{1.01} & \textbf{0.7623} & 0.3228 & 0.8573 & 70.1 & 13.9 & \underline{10.02} & 459s & \$0.40 \\
         & Ours w/o validator & \underline{0.779} & \underline{0.787} & \underline{0.990} & 1.03 & \underline{0.7611} & 0.3240 & 0.8664 & 86.0 & 19.6 & --- & 328s & \$0.20 \\
         & \textbf{Ours (full)} & \textbf{1.000} & \textbf{1.000} & \textbf{1.000} & \textbf{1.00} & 0.7610 & \textbf{0.3243} & \underline{0.8669} & 75.8 & 16.6 & \textbf{2.48} & 251s & \$0.19 \\
        \midrule
        \multirow{3}{*}{\rotatebox[origin=c]{90}{6 Astra}} & Codex & \underline{0.693} & \underline{0.709} & 0.822 & 1.87 & 0.7609 & 0.3237 & 0.8691 & 75.5 & 15.3 & --- & 244s & \$1.21 \\
         & Articraft~\cite{zhou2026articraft} & 0.684 & 0.693 & \underline{0.844} & \underline{1.77} & \underline{0.7690} & \textbf{0.3249} & \textbf{0.8785} & 63.2 & 16.0 & --- & 426s & \$3.00 \\
         & \textbf{Ours (full)} & \textbf{1.000} & \textbf{1.000} & \textbf{1.000} & \textbf{1.00} & \textbf{0.7695} & \underline{0.3243} & \underline{0.8698} & 89.3 & 18.9 & \textbf{2.40} & 267s & \$1.28 \\
        \bottomrule
    \end{tabular*}
\end{table*}

%% file: fig/ablation.tex
\begin{figure*}[t]
  \centering
  \includegraphics[width=\textwidth]{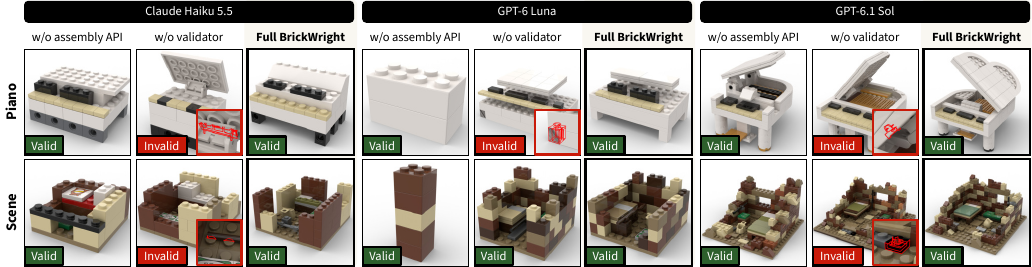}
  \caption{\textbf{Qualitative ablation results.} Red outlines in insets indicate mesh intersections. See \cref{sec:qualitative-prompts} for input text prompts.}
  \label{fig:ablation}
\end{figure*}

%% file: fig/extensions.tex
\begin{figure}[t]
  \centering
  \includegraphics[width=\columnwidth]{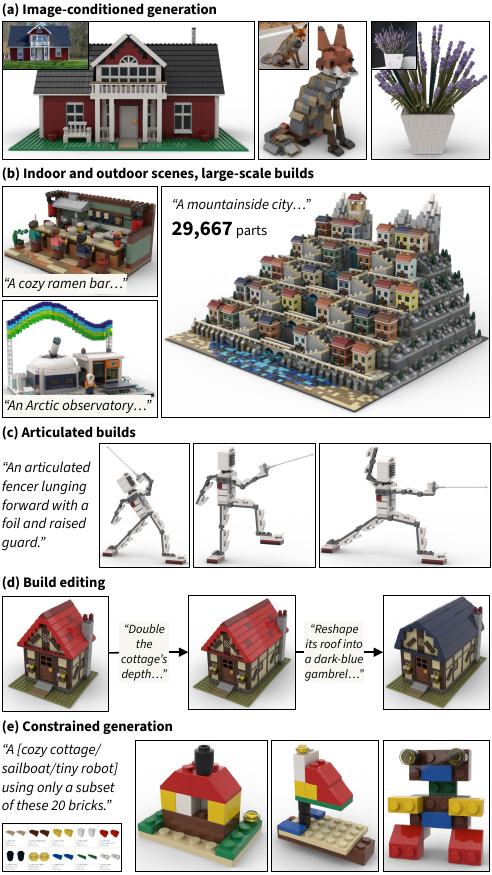}
  \caption{\textbf{Extensions.} BrickWright naturally supports image-conditioned generation, scenes, articulated builds, editing, and constrained generation. All examples are physically valid and connected. See \cref{sec:qualitative-prompts} for full input prompts.}
  \label{fig:extensions}
\end{figure}

%% file: sec/5_experiments.tex
\section{Experiments}

We compare BrickWright against two types of baselines: specially trained text-to-brick models (BrickGPT~\cite{pun2025brickgpt} and BrickNet~\cite{kulits2026bricknet}) and pre-trained agentic methods (Articraft~\cite{zhou2026articraft}, OpenAI's Codex, and Anthropic's Claude Code). We evaluate the physical feasibility and prompt adherence of the output brick builds. Throughout evaluation, we use the BrickNet-512~\cite{kulits2026bricknet} dataset, which contains 512 text prompts created by VLM captioning of human-made brick builds.

\paragraph{Technical details.} To test agents with a range of capabilities, we run experiments with Claude Haiku 5.5 and GPT-6 Luna (lightweight), GPT-6.1 Sol (balanced), and GPT-6 Astra (frontier). All agents use High reasoning. Figure examples use Astra unless otherwise specified. Agents are instructed to create builds with no collisions or floaters, and to ensure the build is connected if it comprises one object. Codex and Claude Code baselines may read, write, and execute arbitrary files in a workspace directory, which initially contains only the LDraw documentation~\cite{ldrawdocumentation} and the library of LDraw parts~\cite{ldrawlibrary}, including part meshes. BrickWright agents are only allowed to read and modify the output \texttt{build.py} and execute text-output-only Python snippets; they have no access to the LDraw documentation or LDraw part library beyond the searchable parts list exposed by BrickLayer. Agentic methods have a budget of \$5 per sample; for samples that reach the budget, we use the latest completed build, whether valid or invalid. Only Luna w/o assembly API (2 samples) and Astra with Codex (1 sample), Articraft (34 samples), and full BrickWright (2 samples) reached this limit.

As Articraft is intended for primitive-based 3D generation, we minimally modify it to use a library of LDraw part meshes rather than mesh primitives. Invalid operations on physical parts (e.g., scaling) are disabled, as are collision checks, which yield false positives for coupled bricks, and watertightness checks, which fail on library meshes. We adjust Articraft's system prompt to reword phrases that do not work with brick builds, such as allowing geometry overlaps.

\paragraph{Metrics.} We report metrics on physical feasibility as defined in \cref{sec:method}; prompt adherence as judged by VQAScore~\cite{lin2024vqascore}, Perception Encoder~\cite{bolya2025perception}, and SigLIP 2~\cite{tschannen2025siglip2}; average brick count per build; and average time and cost per prompt. Validity and connectivity are first evaluated using BrickLayer's validator. Then, to independently verify the metrics' accuracy, we flag potential false positives for review by a human experienced with \LEGO{} design. Potential false positives include collisions resolved by a tested shift of at most 0.2 mm and disconnected components containing parts without connector annotations. Human review increases collision-free rates by 0.2 percentage points for BrickNet and Claude Code (Haiku), 1.4 for Codex (Sol), and 1.0 for Codex (Astra). It reduces mean component counts by 0.13 for Codex (Sol), 0.10 for Codex (Astra), and 0.12 for Articraft; changes for the remaining methods are below 0.02 components per build. Numbers in \cref{tab:main-quantitative} are post-review.

\input{fig/main-qualitative}

\subsection{Baseline comparison}

Compared to vanilla agent baselines, builds produced by BrickWright-equipped agents exhibit greatly improved physical feasibility with broadly comparable token costs~(\cref{tab:main-quantitative}). Impressively, the lightweight model Luna, which has only a 9\% baseline validity rate, achieves 100\% validity when used with BrickWright. This increase in physical validity is not accompanied by a significant decrease in complexity or fidelity; on average, BrickWright agents use more bricks and more unique parts than vanilla agent baselines, while improving prompt adherence for lightweight models (Haiku, Luna) and maintaining it for powerful models (Sol, Astra). These gains hold even with BrickWright’s minimal harness, whereas Codex and Claude Code can read, write, and execute arbitrary files. BrickWright-equipped agents additionally outperform specially trained text-to-brick models in overall feasibility and prompt adherence. The Articraft baseline, which is intended for digital 3D assets, yields comparable prompt adherence but worse physical validity at more than twice the cost of BrickWright. The visual results in \cref{fig:main-qualitative} show how BrickWright reduces physical artifacts for all agent backends and improves the coherence of builds created by lightweight agents. On average, most BrickWright agents use 2--3 \texttt{validate} calls; Luna uses 4--5 (\cref{tab:main-quantitative}).

\subsection{Ablation study}

We ablate BrickLayer's two key components: the assembly API and the validator. In our first ablation, ``w/o assembly API'', agents write a program that creates an LDraw file without using BrickLayer's assembly functions. The \texttt{LegoBuild} abstraction is removed from BrickLayer, leaving only part search and physical validation, and its documentation is replaced by LDraw documentation of similar length. In our second ablation, ``w/o validator'', agents are still told to make their output physically valid but cannot use the \texttt{validate} and \texttt{render} tools or BrickLayer's physical validation functions. They are allowed to run their own validations using \texttt{explore}. We run ablations using Haiku, Luna, and Sol; we exclude Astra due to budget constraints.

BrickLayer's assembly API greatly reduces token usage and makes it easier to create valid builds; when forced to work in LDraw rather than using \texttt{LegoBuild}, agents spend around twice the token cost and three times as many \texttt{validate} calls (\cref{tab:main-quantitative}). Though stronger agents (Sol) still achieve 100\% validity with no assembly API, weaker agents (Luna and Haiku) sometimes give up without finalizing the build, leading to validity scores below 100\%. Without an assembly API, Luna's prompt-similarity scores degrade significantly; it often produces simple structures (\cref{fig:ablation}) and its average part count drops from 27.4 to 7.2. Interestingly, merely using BrickLayer's API \textit{without a provided validator} boosts validity by up to 44 percentage points from Codex/Claude Code baselines! The automatic grid-snapping and connector-mating functions provided by BrickLayer help agents, especially lightweight ones, produce valid builds more easily. However, the validator remains a crucial component; without it, all agents score below 79\% validity (\cref{tab:main-quantitative}).

\subsection{Extensions}

BrickWright works with image conditions, indoor and outdoor scenes, large-scale builds of more than 29,000 parts, articulated objects, build editing by providing a BrickLayer script or LDraw file as input, and special requirements such as being constrained to a given set of bricks (\cref{fig:extensions}). This functionality naturally arises from pre-trained agent capabilities, without needing to update BrickWright's toolkit.

%% file: fig/main-qualitative.tex
\begin{figure*}[p]
  \centering
  \includegraphics[width=\textwidth]{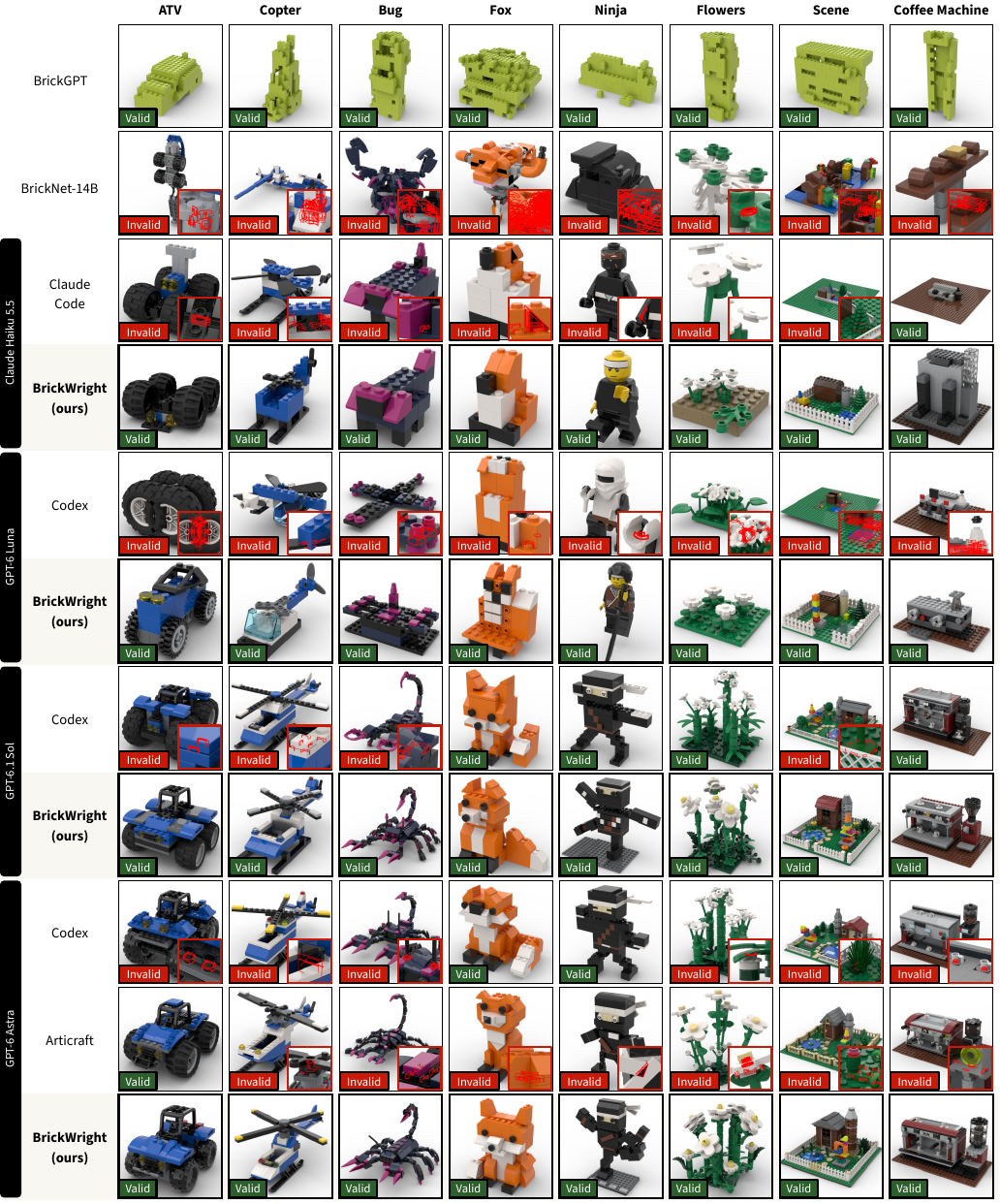}
  \caption{\textbf{Qualitative baseline comparison.} BrickWright improves physical feasibility across all agents and visibly boosts output fidelity for lightweight agents (Haiku, Luna), while outperforming specialized brick generation models~\cite{pun2025brickgpt, kulits2026bricknet}. See \cref{sec:qualitative-prompts} for text prompts and \cref{fig:additional-qualitative} for more results. Red outlines in insets indicate mesh intersections.}
  \label{fig:main-qualitative}
\end{figure*}

%% file: sec/9_conclusion.tex
\section{Discussion}

Frontier coding agents are capable of producing aesthetically pleasing \LEGO{} brick builds from text or image prompts. However, even the most capable agents fail to produce physically feasible builds much of the time. Our framework, BrickWright, closes this gap by introducing a toolkit for brick build creation and validation. Equipped with BrickWright, even lightweight agents achieve 100\% validity on 512 samples tested, while improving prompt adherence at a similar cost to agents without BrickWright. We hope this work highlights the importance of well-written construction APIs and physical validation in improving the performance of LLM agents on building tasks.

\paragraph{Limitations and future work.} Our definition of physical feasibility does not cover valid assembly sequences or stability analysis, which we leave to future work. Though agentic methods outperform specially trained models on brick build generation, they are generally slower and more costly. Effectively distilling the capabilities of powerful agents into smaller, faster models remains an interesting future direction.

\section*{Acknowledgments}
This work is partly supported by the Packard Foundation, Sloan Foundation, and IITP grant funded by the Korean Government (MSIT) (No. RS-2024-00457882, National AI Research Lab Project). Ava Pun is supported by the Meshy Fellowship, Tata Consultancy Services Presidential Fellowship, and NSERC CGRS D. Maxwell Jones is supported by the Rales Fellowship.

%% file: sec/X_suppl.tex
\clearpage
\appendix
\setcounter{page}{1}
\maketitlesupplementary

\section{Additional results}

\input{fig/additional-qualitative} \Cref{tab:valid-at-k-tokens} reports ``valid@k'' metrics: how often agents reach a passing \texttt{validate} call within their first $k$ calls, together with mean token usage. With the assembly API, BrickWright agents pass validation in far fewer calls than when writing LDraw directly; for example, 81\% of GPT-6.1 Sol builds pass within two calls, compared with 37\% without the API.

\Cref{tab:validation-failures} breaks down the reasons for failed \texttt{validate} calls. Physical checks account for most failures (63--85\%), followed by runtime errors (11--28\%) such as selecting a nonexistent connector or specifying an invalid part ID. For GPT agents, most physical failures are due to collisions, while Haiku's main failure mode is connectivity. The more capable an agent, the more its physical failure distribution shifts toward collisions, away from floaters and connectivity.

Additional qualitative comparisons are given in \cref{fig:additional-qualitative}.

\section{System prompt and BrickLayer reference}
\label{sec:system-prompts}

\Cref{lst:brickwright-prompt} shows the system prompt used for BrickWright agents, and \cref{lst:baseline-prompt} shows the user prompt template used for the Codex and Claude Code baselines. The BrickWright prompt also includes the BrickLayer API reference; \cref{lst:legobuild-docs,lst:bricklayer-docs-element,lst:bricklayer-docs-placement,lst:bricklayer-docs-connector,lst:bricklayer-docs-geometry} show excerpts of the reference documenting the \texttt{LegoBuild}, \texttt{Element}, \texttt{Placement}, \texttt{LegoGeometry}, and \texttt{Connector} classes.

\section{Example prompts}
\label{sec:qualitative-prompts}

We provide full prompts for all examples used in the figures.

\paragraph{Teaser (\cref{fig:teaser}).} \textbf{Starry Night:} ``Recreate this painting as a sculptural 3D LEGO scene using varied pieces. No mosaics.'' \textbf{Combine:} ``Match the image as closely as possible.'' \textbf{Seattle Needle:} ``Seattle’s Space Needle, with tapered legs, a saucer observation deck, and its slender spire.'' \textbf{Crab:} ``A poseable crab with eight jointed legs, opening claws, and raised eyestalks.'' \textbf{Harbour:} ``Hong Kong’s Victoria Harbour in 10,000+ parts, with dense skyscrapers and a red-sailed junk.'' \textbf{Spaceship:} ``This is a white and black LEGO spaceship with a blue tinted canopy, featuring a blue-suited minifigure pilot, rear thrusters, and side-mounted weapons.''

\paragraph{Baseline comparison (\cref{fig:main-qualitative}).}
\textbf{ATV:}  ``This is a LEGO build of a blue, grey, and black all-terrain vehicle (ATV) or quad bike, featuring large treaded wheels, a roll cage, and translucent yellow headlights.''
\textbf{Copter:}  ``This is a LEGO build of a blue and white utility helicopter, featuring an open cockpit, black landing skids, and a tail rotor.''
\textbf{Bug:}  ``This is a custom-built LEGO build of a fantastical, insectoid creature, constructed primarily from dark blue and black pieces with vibrant magenta accents for its claws, tail stinger, and decorative elements, showcased from multiple angles to highlight its articulated limbs and unique design.''
\textbf{Fox:}  ``This is a stylized, geometric LEGO build of a sitting fox, constructed primarily from orange bricks with white accents on the chest and snout, black pieces for the paws and nose, and small orange slopes for ears.''
\textbf{Ninja:}  ``This is a blocky, stylized LEGO minifigure of a ninja or martial artist, constructed primarily from black bricks with white and brown accents for its headband and torso, and posed dynamically as if in mid-action.''
\textbf{Flowers:}  ``This LEGO build is a modular, multi-angle representation of a cluster of white flowers with green leaves and stems, constructed from various specialized plant-themed pieces.''
\textbf{Scene:}  ``This LEGO build depicts a small, fenced-in backyard scene featuring a brown shed with an attached silo, a blue pond, green landscaping, and a colorful, abstract sculpture.''
\textbf{Coffee Machine:}  ``This LEGO build is a detailed, dual-grouphead espresso machine with an attached coffee grinder, all mounted on a brown baseplate.''

\input{tab/validation-failures}
\input{tab/valid-at-k-tokens}

\paragraph{Additional baseline comparison (\cref{fig:additional-qualitative}).}
\textbf{Car:}  ``This is a red, open-top LEGO sports car featuring black racing stripes, yellow accents on the front and rear, and translucent orange headlights.''
\textbf{Boat:}  ``This is a LEGO model of a small workboat or tugboat, featuring a red hull, an orange and white cabin with blue windows, a grey deck, and a black articulated crane on the stern.''
\textbf{Train:}  ``This is a small, stylized LEGO model of a yellow and black steam locomotive, featuring a prominent smokestack, a curved boiler, a cab, and simple wheels, presented from multiple angles against a white background.''
\textbf{Snake:}  ``This is a modular, dark green LEGO model of a snake or serpent, constructed from multiple articulated segments that allow it to be posed in various curved and straight configurations.''
\textbf{Squirrel:}  ``This is a LEGO model of a black squirrel, constructed with dark grey and light grey bricks, featuring a large bushy tail, prominent white eyes, and holding two small, round objects in its paws.''
\textbf{Dragon:}  ``This is a custom-built, orange LEGO model of a stylized, blocky creature resembling a dragon or dinosaur, featuring translucent flame elements at its rear and amber pieces for eyes.''
\textbf{Station:}  ``This LEGO model is a white, multi-level medical or scientific station featuring an articulated lamp with a yellow lens, a control panel, and a cabinet with a yellow door.''
\textbf{Garden:}  ``This LEGO model depicts a charming, rotating diorama of a small garden scene, featuring a pumpkin patch with a white fence on one side and a cylindrical well or shed with a ladder on the other, all built upon interconnected green baseplates.''

\paragraph{Ablation study (\cref{fig:ablation}).} \textbf{Piano:}  ``This LEGO model is a compact, white grand piano with tan and black keys, featuring a hinged lid and built from various brick types including technic pieces.''
\textbf{Scene:}  ``This LEGO model depicts a dilapidated, single-room shelter or bunker with crumbling brown and tan walls, containing a simple bed, a rifle, a sack, and a small cache of money.''

\paragraph{Extensions (\cref{fig:extensions}).}
\textbf{Image-conditioned generation:} All three examples use the prompt ``Match the image as closely as possible.''
\textbf{Ramen bar:} ``A cozy ramen bar interior with an open kitchen and diners.''
\textbf{Arctic observatory:} ``An Arctic observatory beneath the aurora.''
\textbf{Mountainside city:} ``A mountainside city in 15,000+ parts, with terraced houses, winding stairs, bridges, and waterfalls.'' \textbf{Build editing:} \textit{Starting prompt}---``A compact timber-framed cottage with a steep red roof.'' \textit{Edit 1}---``Double the cottage’s depth; preserve its width, height, and timber-framed style.'' \textit{Edit 2}---``Reshape its roof into a dark-blue gambrel, keeping the stretched walls and chimney.''

\section{Brick build generation with vanilla agents}
\label{sec:vanilla-analysis}

We give a short analysis of the brick build creation process of baseline GPT-6 Astra, used with Codex. We inspect the Python programs produced and executed by the agent during generation (whether inline or from a file) and classify them as \textit{generators} and/or \textit{validators} by using heuristics that trace the program's abstract syntax tree (AST). If a program calls \texttt{write\_text()} or \texttt{write\_bytes()} on a \texttt{model.ldr} file or \texttt{open}s the file in a mode with write access, we classify it as a generator. If a program reads \texttt{model.ldr} and includes an assertion or error involving values derived from \texttt{model.ldr}, we classify it as a validator. Out of 512 builds created using prompts from BrickNet-512, 471 (92\%) use a generator and 424 (83\%) use a validator.

\clearpage

\input{lst/brickwright_prompt}
\input{lst/baseline_prompt}
\input{lst/legobuild_docs}
\input{lst/element_docs}
\input{lst/placement_docs}
\input{lst/geometry_docs}
\input{lst/connector_docs}

%% file: fig/additional-qualitative.tex
\begin{figure*}[p]
  \centering
  \includegraphics[width=\textwidth,height=0.9\textheight,keepaspectratio]{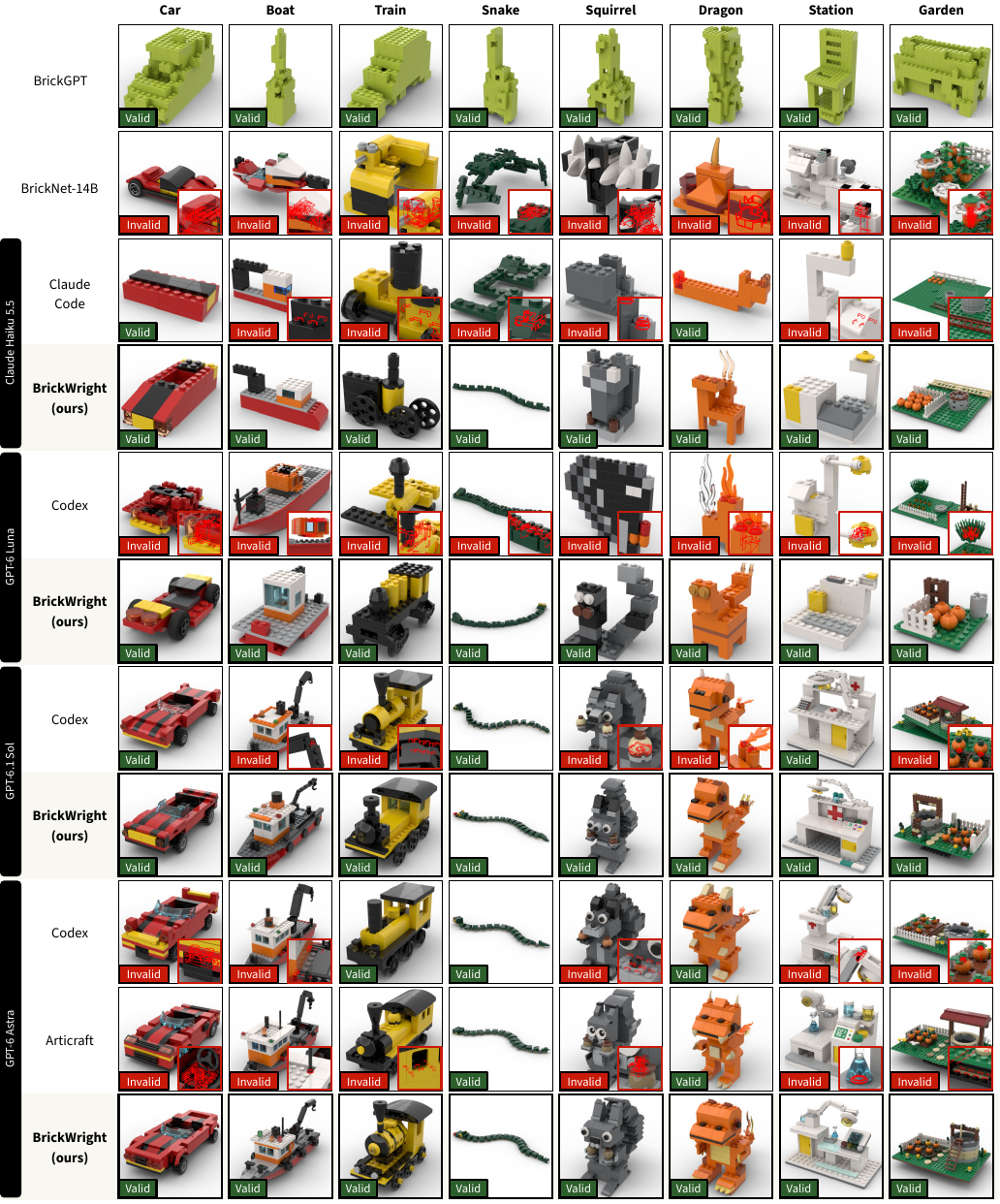}
  \caption{\textbf{Additional qualitative baseline comparison.} See \cref{sec:qualitative-prompts} for text prompts. Red outlines in insets indicate mesh intersections. }
  \label{fig:additional-qualitative}
\end{figure*}

%% file: tab/validation-failures.tex
\begin{table}[t]
    \caption{\textbf{Validation failure types for BrickWright agents on BrickNet-512.} Failure types are given as fractions of failed calls. Runtime errors are exceptions while running \texttt{create\_build()} or \texttt{run\_tests()}; custom test failures pass all physical checks but fail an agent-written test. No calls failed for other reasons. Indented rows break down physical failures as fractions of physical failures; a call can fail several physical checks, so they may sum to $>1$.}
    \label{tab:validation-failures}
    \centering
    \setlength{\tabcolsep}{4pt}
    \begin{tabular}{@{}lcccc@{}}
        \toprule
        & Haiku 5.5 & 6 Luna & 6.1 Sol & 6 Astra \\
        \midrule
        \texttt{validate} calls & 1{,}481 & 2{,}364 & 1{,}269 & 1{,}230 \\
        Failed calls & 805 & 1{,}598 & 477 & 393 \\
        \midrule
        \textbf{Syntax error} & 0.000 & 0.003 & 0.000 & 0.003 \\
        \textbf{Runtime error} & 0.143 & 0.281 & 0.130 & 0.109 \\
        \textbf{Physical test} & 0.820 & 0.635 & 0.813 & 0.847 \\
        \quad Collision & 0.505 & 0.796 & 0.899 & 0.895 \\
        \quad Floater & 0.439 & 0.317 & 0.170 & 0.105 \\
        \quad Connectivity & 0.767 & 0.557 & 0.363 & 0.297 \\
        \textbf{Custom test} & 0.037 & 0.081 & 0.057 & 0.041 \\
        \bottomrule
    \end{tabular}
\end{table}

%% file: tab/valid-at-k-tokens.tex
\begin{table*}[t]
    \caption{\textbf{Validation success and token usage on BrickNet-512.} valid@$k$ is the fraction of builds whose first $k$ \texttt{validate} calls include at least one passing call (all physical checks and custom tests pass); it applies only to methods with access to \texttt{validate}. Tokens are mean thousands of tokens per prompt; total tokens are the sum of input tokens (including cached input) and output tokens (including reasoning). Bold marks the best valid@$k$ within each agent group.}
    \label{tab:valid-at-k-tokens}
    \centering
    \setlength{\tabcolsep}{2pt}
    \begin{tabular*}{\textwidth}{@{}cl@{\hspace{3pt}\extracolsep{\fill}}*{8}{c}@{}}
        \toprule
        & & \multicolumn{5}{c}{valid@$k$$\uparrow$} & \multicolumn{3}{c}{Tokens (K)$\downarrow$} \\
        \cmidrule(lr){3-7} \cmidrule(l){8-10}
        \multicolumn{2}{@{}l}{Method} & $k{=}1$ & $k{=}2$ & $k{=}3$ & $k{=}4$ & $k{=}5$ & Total & Input & Output \\
        \midrule
        \multirow{4}{*}{\rotatebox[origin=c]{90}{Haiku 5.5}} & Claude Code & --- & --- & --- & --- & --- & 603.2 & 570.0 & 33.2 \\
         & Ours w/o assembly API & 0.143 & 0.324 & 0.439 & 0.527 & 0.613 & 1{,}900.7 & 1{,}852.2 & 48.5 \\
         & Ours w/o validator & --- & --- & --- & --- & --- & 562.2 & 540.8 & 21.3 \\
         & \textbf{Ours (full)} & \textbf{0.320} & \textbf{0.637} & \textbf{0.801} & \textbf{0.908} & \textbf{0.959} & 846.4 & 821.7 & 24.7 \\
        \midrule
        \multirow{4}{*}{\rotatebox[origin=c]{90}{6 Luna}} & Codex & --- & --- & --- & --- & --- & 495.5 & 482.6 & 12.9 \\
         & Ours w/o assembly API & 0.217 & 0.334 & 0.482 & 0.596 & 0.678 & 3{,}324.8 & 3{,}288.5 & 36.4 \\
         & Ours w/o validator & --- & --- & --- & --- & --- & 719.9 & 678.6 & 41.3 \\
         & \textbf{Ours (full)} & \textbf{0.268} & \textbf{0.500} & \textbf{0.646} & \textbf{0.750} & \textbf{0.814} & 1{,}282.0 & 1{,}249.8 & 32.2 \\
        \midrule
        \multirow{4}{*}{\rotatebox[origin=c]{90}{6.1 Sol}} & Codex & --- & --- & --- & --- & --- & 654.3 & 644.6 & 9.6 \\
         & Ours w/o assembly API & 0.164 & 0.369 & 0.535 & 0.658 & 0.732 & 1{,}825.0 & 1{,}810.6 & 14.4 \\
         & Ours w/o validator & --- & --- & --- & --- & --- & 334.7 & 321.3 & 13.4 \\
         & \textbf{Ours (full)} & \textbf{0.475} & \textbf{0.807} & \textbf{0.932} & \textbf{0.977} & \textbf{0.992} & 528.3 & 519.0 & 9.4 \\
        \midrule
        \multirow{3}{*}{\rotatebox[origin=c]{90}{6 Astra}} & Codex & --- & --- & --- & --- & --- & 458.5 & 452.1 & 6.3 \\
         & Articraft~\cite{zhou2026articraft} & --- & --- & --- & --- & --- & 1{,}540.2 & 1{,}528.0 & 12.1 \\
         & \textbf{Ours (full)} & \textbf{0.545} & \textbf{0.879} & \textbf{0.953} & \textbf{0.984} & \textbf{0.996} & 544.5 & 533.5 & 10.9 \\
        \bottomrule
    \end{tabular*}
\end{table*}

%% file: lst/brickwright_prompt.tex
\begin{lstlisting}[
    style=bricklayer,
    language={},
    breakindent=0pt,
    float=*t,
    caption={\textbf{BrickWright system prompt.} \texttt{\{docs\}} is replaced by tool documentation (e.g. for \texttt{validate}) and the BrickLayer reference.},
    label={lst:brickwright-prompt}
]
# BrickWright Instructions

You are BrickWright, an expert LEGO builder. Edit `build.py` so `create_build()` returns a `LegoBuild` faithful to the request and `run_tests()` returns its `TestContext` report. The BrickLayer API reference below describes construction, coordinates, connectors, parts, and colors; use it to resolve API questions.

The finished build must meet these requirements:

- **No collisions:** parts may touch and mate, but must not interpenetrate.
- **No floating parts:** every part must reach the ground through mated connections or supporting surface contacts. Resting on tiles, ledges, and slopes counts; side contact alone does not. Sliding and tipping are not checked.
- **Connected:** one connected assembly, unless the request calls for separate pieces. Face contact alone does not connect parts. Use compatible connectors: studs/antistuds, pins/holes, axles, clips/bars, ball joints, and others. Stagger ordinary brick layers or bridge adjacent pieces with a shared plate.
- **Faithful:** preserve the requested identity, proportions, features, and detail at a believable LEGO scale. Do not remove requested features merely to pass checks.

## Checks in build.py

The harness independently enforces no floating parts and no collisions on the built model. Connectivity and request-specific checks come from `run_tests()`. Keep `check_no_floating()`, `check_no_collisions()`, and `check_connected()` in `run_tests()`. Change `expected_components` only when the request genuinely calls for multiple assemblies. Add `tests.check(name, ok, detail)` for measurable requirements; inspect failures and repair the build rather than weakening those requirements.

```python
import bricklayer as bl
from brickwright import TestContext

def create_build():
    build = bl.LegoBuild()
    build.add(bl.Element("3001", bl.Color.RED))
    return build

def run_tests():
    build = create_build()
    tests = TestContext(build)
    tests.check_no_floating()
    tests.check_no_collisions()
    tests.check_connected(expected_components=1)
    # Add checks for this request, e.g. tests.check("over 100 parts", build.brick_count > 100).
    return tests.report()
```

For articulated builds, declare intended movement with `MotionRange` on the relevant connections. Optionally add `tests.check_motion(samples=5)` to sample the annotated ranges for collisions, floating parts, and connector disengagement. Other connections stay rigid, which can produce false positives if they would also move in reality. This check is optional and covers sampled poses, not every intermediate position. `build.check_motion()` exposes its detailed report for investigation. Static validity checks do not test motion.

## Working method

1. Plan the defining features, rough size, support structure, and specialized parts.
2. Look up catalog parts before using remembered IDs. Use `Element(part_id, color)` as the normal constructor; size helpers are conveniences. Check their listed argument combinations: dimensions are width-first, and changing orientation requires rotation, not swapping dimensions to an unsupported size.
3. Build in manageable increments. Validate, read each failure's detail, and repair it. Use helpers and sub-models for repeated or meaningful assemblies.
4. When rendering is available, inspect the build while working and after the final edit. Physical validity alone does not establish visual fidelity.
5. Run the checks on the finished source. Keep `create_build()` deterministic and free of file writes so checks, previews, and exports represent the same build.

{docs}
\end{lstlisting}

%% file: lst/baseline_prompt.tex
\begin{lstlisting}[
    style=bricklayer,
    language={},
    breakindent=0pt,
    float=*t,
    caption={\textbf{Codex/Claude Code baseline prompt.} \texttt{\{caption\}} is replaced with the text prompt.},
    label={lst:baseline-prompt}
]
Build a LEGO model matching the following description. Save the complete LDraw code in model.ldr in the current workspace.

The model must be an assembly of real, existing LEGO parts from the LDraw parts catalogue, referenced by their exact .dat filenames. Use only type-0 comments and metadata and type-1 part or submodel references. Do not emit type-2, type-3, type-4, or type-5 geometry, invent part IDs, or define custom part geometry. If you use MPD submodels, embed them in model.ldr and build each entirely from catalogued parts. Every leaf reference must resolve to a real catalogue part.

Avoid colliding or floating parts. If the model depicts a single object, all its parts must form one connected component through valid LEGO connections.

Description:
{caption}
\end{lstlisting}

%% file: lst/legobuild_docs.tex
\begin{lstlisting}[
    style=bricklayer,
    language={},
    breakindent=0pt,
    float=*t,
    caption={\textbf{\texttt{LegoBuild} documentation.} A \texttt{LegoBuild} represents a multi-part assembly containing placed elements or sub-builds. It supports adding/removing placements and validating the build's physical feasibility. Excerpt from the BrickLayer API reference included in the BrickWright system prompt. Continued on the next two pages.},
    label={lst:legobuild-docs}
]
#### class `LegoBuild(LegoGeometry)`

`LegoBuild(placements: Iterable[Placement] = (), names: Mapping[str, Placement] | None = None, steps: Iterable[int] | None = None, *, joints: Iterable[Joint] = ()) -> None`

A build or reusable sub-assembly, edited in place.

Validity queries update automatically after edits, including edits to nested models. `is_valid` checks collisions and ground support; `is_connected` checks whether every part joins through compatible connectors.

**Properties**

- `bounding_box: tuple[Position, Position]` -- The physical bounding box spanning every part's, in the model's frame.

  A sub-model holding no parts spans nothing, so a model with no parts returns ``(Position.ZERO, Position.ZERO)``.
- `brick_count: int` -- The total number of bricks in the model.
- `colliding_placements: tuple[tuple[Placement[Element], Placement[Element]], ...]` -- Pairs of placements whose collision volumes overlap.
- `connected_components: tuple[tuple[Placement[Element], ...], ...]` -- The placements grouped by connected component.

  Each group holds the parts joined to one another through couplings, with the groups and the parts within them in placement order. An empty model has no components.
- `connectors: frozenset[Connector]` -- Unoccupied connector points in this model's frame.

  A connector disappears here when it mates with another part, including across nested sub-model boundaries. Covering studs can therefore change the model's alignment box and how subsequent parts stack on it.
- `flattened_joints: tuple[Joint, ...]` -- All joint occurrences, with endpoint paths rooted in this model.

  A repeated sub-model contributes distinct records for each occurrence. Use these records as keys to ``posed`` when moving a nested joint.
- `flattened_placements: tuple[Placement[Element], ...]` -- Every element placement, flattened into this model's own frame.

  Flattening reaches through nested sub-models, so each leaf holds an :class:`Element`.
- `floating_placements: tuple[Placement[Element], ...]` -- Parts without a path to the ground through connectors or contact.

  The lowest physical extent defines the ground, as in `translate_to_ground`. The model's poses remain unchanged, and returned placements are the original leaves.

  Mated connectors transmit support in both directions. Surface contacts transmit it from the supporting part to the part resting on it, including on tiles, elevated ledges, and slopes. Sliding and tipping are not checked.
- `joints: tuple[Joint, ...]` -- Connections recorded in this model; nested records stay in their owner.

  Removing either endpoint, including a nested leaf, deletes its record.
- `names: Mapping[str, Placement]` -- The model's named placements, keyed by the name each was registered under.

  A live read-only view: names appear as parts are added.
- `num_connected_components: int` -- The number of connected components of the model.
- `placements: tuple[Placement, ...]` -- Every top-level placement in the model, in insertion order.
\end{lstlisting}

\begin{lstlisting}[
    style=bricklayer,
    language={},
    breakindent=0pt,
    float=*t,
    title={\textbf{\texttt{LegoBuild} documentation} (continued).}
]
**Methods**

- `add(content: Element | LegoBuild, position: Position | tuple[float, float, float] = Position(0, 0, 0), rotation: Rotation = Rotation.DEG_0, anchor: Anchor = Anchor.FRONT_RIGHT, *, relative_to: NameRef | Placement | None = None, target_anchor: Anchor = Anchor.FRONT_RIGHT, name: str | None = None, index: int | None = None) -> Placement` -- Place content on the grid; return its placement.

  `position` is an absolute `(x, y, z)` or a delta from `relative_to` (a name, name path, or placement). `rotation` is the content's own rotation, not relative to the reference. Positions require integer x/y and z in whole plates.

  `anchor` selects the point on the content's rotated alignment footprint; `target_anchor` selects the reference point when using `relative_to`. The default is the front-right minimum corner. An even extent's center floors toward that corner: `Anchor.CENTER` on a 4x2 footprint is `(1, 0)` studs from it.

  `name` registers the result; `index` inserts instead of appending. Content is held by reference. A concrete reference is read in this model's frame, even if from another model or removed; it is not added automatically.

  Raises `ValueError` for off-grid positions, `KeyError` for missing names, and `IndexError` for an out-of-range insertion.
- `add_raw(content: Element | LegoBuild, pose: Pose = Pose(rotation=Mat3.IDENTITY, translation=Position(0, 0, 0)), *, name: str | None = None, index: int | None = None) -> Placement` -- Place content at an arbitrary pose without grid checks or re-anchoring.

  `pose` maps the content's local frame into this model. An element's origin is its alignment box's bottom-center; a sub-model retains its own frame. Content is shared by reference. `name` registers the placement and `index` inserts instead of appending. Invalid insertion indices raise `IndexError`.
- `check_motion(*, samples: int = 5, max_samples: int = 10000) -> MotionReport` -- Optionally check collisions, floating, and engagement across joint ranges.

  Samples the Cartesian product, including zero and both limits on every coordinate. Unannotated connections remain rigid, which can flag motions possible with additional flexibility. A clear report covers only the samples. Exceeding the sample budget or encountering a motion loop raises.
- `connect(content: Element | LegoBuild, connector: Connector | ConnectorType | str, to: Connector, *, mounting: Mounting | None = None, motion: MotionRange | None = None, target_rotation: Mat3 | Rotation | None = None, target_position: Position | tuple[float, float, float] | None = None, name: str | None = None, index: int | None = None) -> Placement` -- Place content by mating a local connector to a world target.

  ``target_rotation`` and ``target_position`` fit the pose; ``mounting`` then adds clocking, insertion, flip, or ball swing. Motion is relative to this pose.

  A target in this model records a static joint. Set ``motion`` to a ``MotionRange`` matching the joint's ``kind`` to enable movement.

  Both connectors must identify unique leaf occurrences. A geometric target outside this model may still position static content, but cannot declare motion. No placement or joint is recorded if validation fails.
\end{lstlisting}

\begin{lstlisting}[
    style=bricklayer,
    language={},
    breakindent=0pt,
    float=*t,
    title={\textbf{\texttt{LegoBuild} documentation} (continued).}
]
- `copy() -> LegoBuild` -- A shallow copy holding the same placements under the same names.

  Placements and nested sub-models are shared. Build-step markers carry over. Adding to either outer model leaves the other alone; editing a shared sub-model affects both. Use :meth:`deepcopy` to isolate those edits.
- `deepcopy() -> LegoBuild` -- Copy this model and nested sub-models, isolating subsequent edits.

  Names, poses, and build steps are preserved. Repeated sub-models stay shared within the copy, but no mutable content is shared with the original.
- `extend(other: LegoBuild, position: Position | tuple[float, float, float] = Position(0, 0, 0), rotation: Rotation = Rotation.DEG_0, anchor: Anchor = Anchor.FRONT_RIGHT) -> tuple[Placement, ...]` -- Splice `other`'s current top-level placements into this model.

  The resulting geometry matches `add(other, position, rotation, anchor)`, but without a wrapper sub-model or its names. Returns the new placements in order. Later additions to `other` do not carry over; nested content remains shared. Raises `ValueError` for off-grid positions or cyclic nesting.
- `is_connected() -> bool` -- Whether the model has a single connected component.
- `is_valid() -> bool` -- Whether the model has no collisions or floating parts.
- `joint(child: Placement) -> Joint` -- The locally recorded joint that placed ``child`` using ``connect``.
- `list_elements() -> Counter[Element]` -- How many of each element -- part and color -- the model uses.
- `name_of(placement: Placement) -> str | None` -- The name ``placement`` was registered under, or ``None`` if unnamed.
- `named(name: NameRef) -> Placement` -- Look up a name or nested path such as `['tower', 'roof']`.

  Returns a placement in this model's frame, suitable for `relative_to` or `on`. That placement remains usable after removal, but looking up its removed name raises `KeyError`. An empty path or traversal through an element raises `ValueError`.
- `posed(positions: Mapping[Joint, JointCoordinates]) -> LegoBuild` -- A flat geometry snapshot at joint coordinates, leaving this build intact.

  Keys come from ``joints`` or ``flattened_joints``. Unspecified positions stay at zero. Static connections and attached rigid assemblies follow the moving parts. The snapshot has no joint records; evaluate every pose from the original build. Closed motion loops raise ``ValueError``.
- `remove(target: int | str | Placement) -> Placement` -- Remove and return a top-level placement by index, name, or identity.

  Negative indices count from the end. Repeated placements lose their first occurrence; the name is freed after the last. Build steps track the removal. Nested name paths are not accepted: edit the sub-model itself, which affects all models sharing it. Missing indices/names raise `IndexError`/`KeyError`; unknown placements raise `ValueError` and unsupported target types `TypeError`.
- `set_motion(joint: Joint | Placement, motion: MotionRange | None) -> Joint` -- Replace a local joint's range; ``None`` makes it static. Return the record.
- `stack(content: Element | LegoBuild, on: NameRef | Placement, offset: Position | tuple[float, float, float] = Position(0, 0, 0), rotation: Rotation = Rotation.DEG_0, anchor: Anchor = Anchor.FRONT_RIGHT, target_anchor: Anchor = Anchor.FRONT_RIGHT, *, name: str | None = None, index: int | None = None) -> Placement` -- Place content on top of `on` (a name, name path, or placement).

  The new part's alignment-box bottom rests at the reference's alignment-box top. `offset` shifts from there in studs/bricks and must be on the grid. Use `anchor=Anchor.CENTER, target_anchor=Anchor.CENTER` to center footprints; anchors, absolute rotation, shared content, naming, and insertion behave as in `add`.

  A concrete reference is read in this model's frame even if removed or from another model. Later changes to that reference do not move this placement.
\end{lstlisting}

%% file: lst/element_docs.tex
\begin{lstlisting}[
    style=bricklayer,
    language={},
    breakindent=0pt,
    float=*t,
    caption={\textbf{\texttt{Element} documentation.} An \texttt{Element} represents a single colored part. Excerpt from the BrickLayer API reference.},
    label={lst:bricklayer-docs-element}
]
#### class `Element(LegoGeometry)`

`Element(ldraw_id: str, color: Color | LDrawColor) -> None`

A catalogued part in a color, constructed as `Element(part_id, color)`.

Elements are immutable; matching part ids and colors compare equal.

**Properties**

- `bounding_box: tuple[Position, Position]` -- Physically-measured `(min, max)` box in the local frame.
- `color: Color | LDrawColor` -- The element's color.
- `connectors: frozenset[Connector]` -- Points where a connection can be made, in the local frame.
- `ldraw_id: str` -- The part's LDraw id.
- `spec: PartSpec` -- The part's catalog identity, resolved from its id.
\end{lstlisting}

%% file: lst/placement_docs.tex
\begin{lstlisting}[
    style=bricklayer,
    language={},
    breakindent=0pt,
    float=*t,
    caption={\textbf{\texttt{Placement} documentation.} A \texttt{Placement} pairs an element or nested build with a rigid transformation. Excerpt from the BrickLayer API reference.},
    label={lst:bricklayer-docs-placement}
]
#### class `Placement(LegoGeometry)`

An element or sub-model placed in a containing model.

`content` and `pose` are fixed. Geometry is expressed in the containing model's frame and updates if the content is a mutable sub-model. Placements compare by identity; two placements can hold identical elements at different positions.

**Properties**

- `bounding_box: tuple[Position, Position]` -- The content's physical bounding box, in world coordinates.
- `connectors: frozenset[Connector]` -- The content's connector points, in world coordinates.
- `pose: Pose` -- Maps the content's local frame into the model's.

**Methods**

- `to_world(p: Position | tuple[float, float, float]) -> Position` -- Map a position from the content's local frame to world coordinates.
\end{lstlisting}

%% file: lst/geometry_docs.tex
\begin{lstlisting}[
    style=bricklayer,
    language={},
    breakindent=0pt,
    float=*t,
    caption={\textbf{\texttt{LegoGeometry} documentation.} \texttt{LegoGeometry} is the base class of \texttt{Element} and \texttt{LegoBuild}, exposing dimension and connector information. Excerpt from the BrickLayer API reference.},
    label={lst:bricklayer-docs-geometry}
]
#### class `LegoGeometry`

Geometry shared by elements, placements, and models.

Elements use local coordinates; placements use the containing model's frame. Models use the frame they were built in. Measurements update when content changes.

**Properties**

- `alignment_box: tuple[Position, Position]` -- The `(min, max)` box used for grid placement and stacking.

  Grid-aligned connectors contribute 1x1 stud footprints and determine the bottom. The highest upward stud sets the top; without one (e.g. a tile), the physical bounding box sets the top. Models use only their unoccupied connectors.

  Without grid-marked connectors, it is derived from the physical bounding box. Width/depth are whole studs and height is in whole plates; thin parts can have zero alignment extent. Inspect the box rather than guessing offsets.
- `alignment_depth: int` -- Extent along Y (studs) of the alignment box. Always an integer.
- `alignment_height: float` -- Extent along Z (bricks) of the alignment box. Always an integer multiple of PLATE.
- `alignment_width: int` -- Extent along X (studs) of the alignment box. Always an integer.
- `center: Position` -- The midpoint of the bounding box, in BrickLayer coordinates.

  Empty geometry returns ``Position(0, 0, 0)``.
- `depth: int | float` -- Extent along Y (studs) of the bounding box.
- `height: int | float` -- Extent along Z (bricks) of the bounding box.
- `width: int | float` -- Extent along X (studs) of the bounding box.

**Methods**

- `find_connector(connector_type: ConnectorType | str | Iterable[ConnectorType | str] | None = None, toward: Position | tuple[float, float, float] | Direction | None = None, *, facing: Direction | Vec3 | None = None) -> Connector` -- Find one connector using the filters and ordering of :meth:`find_connectors`.

  With ``toward``, return the first ranked match. Otherwise require exactly one match. Raises ``ValueError`` if none match or the choice is ambiguous. ``connect`` takes a world-frame target connector from a placement.
- `find_connectors(connector_type: ConnectorType | str | Iterable[ConnectorType | str] | None = None, toward: Position | tuple[float, float, float] | Direction | None = None, *, facing: Direction | Vec3 | None = None, count: int | None = None) -> list[Connector]` -- The geometry's connectors, filtered and ordered by the given criteria.

  ``connector_type`` filters types; ``facing`` filters axis direction. ``toward`` orders nearest a position or furthest in a direction; ``None`` uses a stable geometric order. ``count`` caps the results (default: all). Coordinates are local for an element, world for a placement.
- `rotated_footprint(rotation: Rotation) -> tuple[int, int]` -- The alignment box's ``(width, depth)`` in studs after ``rotation``.

  An odd quarter turn swaps the two; a half turn leaves them as they are.
\end{lstlisting}

%% file: lst/connector_docs.tex
\begin{lstlisting}[
    style=bricklayer,
    language={},
    breakindent=0pt,
    float=*t,
    caption={\textbf{\texttt{Connector} documentation.} A \texttt{Connector} is a typed mating point on an element or build. Connectors come with a local frame and length; compatibility and relative frame alignment determine valid connections. Excerpt from the BrickLayer API reference.},
    label={lst:bricklayer-docs-connector}
]
#### class `Connector`

A mating point obtained from a part or placement's connectors.

`position` is in the queried geometry's frame. `axis` is its mating direction; `side` fixes its roll. Both directions use uniform stud units. `length` is the sliding reach in studs on either side of the point.

**Properties**

- `axis: Vec3`
- `length: Scalar`
- `position: Position`
- `side: Vec3`
- `type: ConnectorType`
\end{lstlisting}